\documentclass{article}

\usepackage{PRIMEarxiv}

\usepackage[utf8]{inputenc} % allow utf-8 input
\usepackage[T1]{fontenc}    % use 8-bit T1 fonts
\usepackage{hyperref}       % hyperlinks
\usepackage{url}            % simple URL typesetting
\usepackage{booktabs}       % professional-quality tables
\usepackage{amsfonts}       % blackboard math symbols
\usepackage{amsmath}
\usepackage{nicefrac}       % compact symbols for 1/2, etc.
\usepackage{microtype}      % microtypography
\usepackage{lipsum}
\usepackage{fancyhdr}       % header
\usepackage{graphicx}       % graphics
\usepackage{cleveref}
\graphicspath{{media/}}     % organize your images and other figures under media/ folder

\usepackage{xcolor}

\title{Learning Sparse Nonlinear Dynamics \\ via Mixed-Integer Optimization
}

\author{
  Dimitris Bertsimas \\
  Sloan School of Management and Operations Research Center\\
  Massachusetts Institute of Technology\\
  \texttt{dbertsim@mit.edu} \\
\and
  {\bf Wes Gurnee} \\
  Operations Research Center \\
  Massachusetts Institute of Technology\\
  \texttt{wesg@mit.edu} \\
}

\begin{document}
\maketitle

\begin{abstract}
% Discovering governing equations is important problem
% Lots of work to do this powered by heuristic sparse 
Discovering governing equations of complex dynamical systems directly from data is a central problem in scientific machine learning. In recent years, the sparse identification of nonlinear dynamics (SINDy) framework, powered by heuristic sparse regression methods, has become a dominant tool for learning parsimonious models. We propose an exact formulation of the SINDy problem using mixed-integer optimization (MIO) to solve the sparsity constrained regression problem to provable optimality in seconds. On a large number of canonical ordinary and partial differential equations, we illustrate the dramatic improvement of our approach in accurate model discovery while being more sample efficient, robust to noise, and flexible in accommodating physical constraints.
\end{abstract}

% keywords can be removed
\keywords{Sparse Regression \and Nonlinear Dynamics \and Optimization}

\section{Introduction}

Advances in machine learning (ML) combined with the exponential growth of data and computing power are enabling new paradigms of data-driven science and engineering. In particular, emerging techniques for learning dynamic patterns directly from data are poised to revamp fields where data is abundant but traditional static analysis methods have failed to generate useful models. While accurate dynamical models are important, the ultimate goal is to advance scientific understanding by discovering interpretable models that are as simple as possible, but no simpler.

The modern era of data-driven system discovery started in earnest with the work of \cite{bongard2007automated, schmidt2009distilling} on symbolic regression. Since then, probabilistic methods \cite{pan2015sparse, zhang2018robust, fuentes2019efficient, rudy2021sparse} and deep neural networks \cite{bar2019learning, champion2019data, lusch2018deep, raissi2017physics, raissi2018multistep, wehmeyer2018time, yeung2019learning} have proven to be effective tools for modeling high-dimensional complex dynamical systems.
However, it is the seminal work of \cite{brunton_discovering_2016} on the Sparse Identification of Nonlinear Dynamics (SINDy) framework that serves as the foundation for our approach. SINDy casts system identification as a sparse regression problem over a large set of nonlinear library functions to find the fewest active terms which accurately reconstruct the system dynamics. Such a technique is especially useful for finding highly interpretable models, and performs well even with limited training data (in particular, much less than what a neural network would require). Its success has inspired a large number of extensions and variants tailored for more specific problems \cite{brunton2016sparse, rudy2017data, kaiser_sparse_2018, kaheman_sindy-pi_2020}. 

% Existing optimization paragraph
The enabling technology underlying all of these methods is the optimization algorithm that selects and fits the best set of library terms to reconstruct the dynamics. The original SINDy paper \cite{brunton_discovering_2016} used the sequential threshold least squares algorithm which finds a sparse solution by iteratively fitting a least squares regression on the candidate library and removing terms with coefficients below a specified threshold. However, thresholding is problematic for recovering small model coefficients and does not easily allow adding additional structure on the coefficients (e.g., enforcing physical symmetries). These limitations motivated the development of algorithms with different sparse regularizers that could incorporate constraints such as the sparse relaxed regularized regression (SR3) \cite{zheng2018unified, champion2020unified} and Conditional Gradients based approaches \cite{carderera_cindy_2021}. There is also a long history of greedy algorithms and convex relaxations designed to solve sparse regression and related subset selection problem \cite{pati1993orthogonal, tibshirani1996regression, tillmann2021cardinality}. For system identification problems, we claim that such relaxations and heuristics are unnecessary, and ultimately, inadequate.

Concurrent with the developments in system identification and increase in scientific data, advances in hardware and software have led to over a one trillion times speed up in mixed-integer optimization (MIO) solvers since 1990 \cite{bixby2002solving, achterberg2013mixed, kronqvist2019review}. This fact necessitates researchers revisit their preconceptions about the (in)tractability of MIO in machine learning contexts. Indeed, there has been substantial work on viewing the full slate of classical machine learning algorithms under a modern optimization lens \cite{bertsimas2019machine}, often yielding state-of-the-art results with practical computational budgets. Most relevant to system identification is the recent progress on high-dimensional sparse regression which solves the NP-hard feature selection problem exactly \cite{bertsimas2016best, bertsimas2020sparse, bertsimas2020sparse_jean, bertsimas_scalable_2020, hazimeh_sparse_2021-1, bertsimas2022backbone, tillmann2021cardinality}. In addition to superior performance, these modern formulations inherit the full generality of MIO, empowering domain experts with a rich modeling language to express a vast range of model desiderata as arbitrary linear, quadratic, and semidefinite constraints on both the coefficients and sparsity structure.

% The objective of this work is to bridge the gap between the system identification and discrete optimization literatures and demonstrate the effectiveness of learning sparse nonlinear dynamics via MIO. We begin by reviewing MIO for sparse regression and then adapt a formulation utilizing specially ordered sets to the basic SINDy framework and its relevant extensions. We then systematically illustrate the contrast in performance between heuristic and MIO sparse regression methods on a wide range of canonical dynamical systems, including both ordinary and partial differential equations. In particular, we focus on accurate support recovery under both low data and high noise regimes, as interpretability and statistical efficiency are the main strengths of SINDy over deep learning approaches. We also illustrate the ability to encode physical constraints and compare the computational efficiency of different algorithms. Our results firmly establish that MIO is fully tractable and achieves state-of-the-art performance in system identification problems.

% Alternative
The objective of this work is to bridge the gap between the system identification and discrete optimization literatures and demonstrate the effectiveness of learning sparse nonlinear dynamics via MIO. We begin by reviewing MIO for sparse regression and then adapt a formulation utilizing specially ordered sets to the basic SINDy framework and its relevant extensions. We then systematically illustrate the contrast in performance between heuristic and MIO sparse regression (MIOSR) methods on a wide range of canonical dynamical systems, including both ordinary and partial differential equations. Our main contribution is the MIOSR algorithm for SINDy for which we establish the following results:
\begin{enumerate}
    \item {\bf Tractable and provably optimal.} MIOSR terminates when the objective of the incumbent solution matches the dual lower bound. That is, when the gap between the objective upper bound and lower bound vanishes, yielding both an optimal solution and a proof of optimality. Despite solving the NP-hard subset selection problem exactly, in Section~\ref{sec:comp_eff} we show that the extra computational cost of MIOSR is minimal, and scales favorably with additional data and compute.
    \item {\bf Sample efficient and noise robust.} This theoretical optimality buys practical performance in more challenging statistical regimes. In particular, we study the low data limit in Section~\ref{sec:sample_eff} and the high noise setting in Section~\ref{sec:robustness} where we find MIOSR outperforms heuristic methods, especially in learning the true sparse form of the dynamics.
    \item {\bf Customizable.} Due to the flexibility of MIO as a modeling framework, MIOSR can be endlessly customized to impose additional structure on the learning problem to further improve sample efficiency, enforce physically realistic models, or incorporate other domain tailored model requirements. We discuss this flexibility and a few relevant extensions in Section~\ref{subsec:extensions} and then demonstrate the benefits of incorporating known physics as constraints in Section~\ref{sec:physcon}.
    \item {\bf Consistent interface.} We provide an implementation of our algorithm which adheres to the PySINDy interface \cite{de_silva_pysindy_2020, Kaptanoglu2022}, both computationally and conceptually. Therefore, our algorithm is compatible with other advancements in the SINDy framework (e.g., preprocessing, library construction, outer loop algorithms) and seamlessly integrates into existing tools and workflows.
\end{enumerate}

\section{Methods}

We begin by reviewing the SINDy problem, its extension to partial differential equations (PDEs), and the weak form of SINDy for noisy data. We then discuss the most appropriate MIO formulations for sparse regression that solve the SINDy problem and some relevant extensions.
\subsection{SINDy}
The original SINDy framework \cite{brunton_discovering_2016} was designed to recover systems of ordinary differential equations (ODEs) of the form 
\begin{equation}\label{eq:system}
    \frac{d}{dt}x(t) = f(x(t))
\end{equation}
where $x(t) \in \mathbb{R}^d$ is the state of the system at time $t$, and the function $f(x(t))$ encodes the dynamics of the system. Implicitly, $f(\cdot)$ is also assumed to be sparse. This is justified because most physical systems are known to have sparse dynamics when represented in suitable coordinates. Sparsity also acts as a natural and effective regularizer \cite{kutz2022parsimony} while yielding more interpretable models.

Given $n$ measurements of a system of interest, the three required inputs are a state-time matrix ${\bf X} \in \mathbb{R}^{n \times d}$ where $x_{ij}$ is the state of system variable $j$ at time $i$, the measured or numerically approximated time derivatives of the state variables ${\bf \dot{X}}$, and a candidate library of nonlinear functions $\bf{\Theta(X)} = [ \theta_1 (\mathbf{X}), \dots, \theta_L (\mathbf{X}) ]$. As an example, we could consider second order polynomials $\theta_\ell(\mathbf{X}) = \mathbf{X}^2$ which would yield $d(d-1)/2$ candidate terms of elementwise products for every pair $\mathbf{X}_i \odot \mathbf{X}_j$.  In all of our experiments, we will use a library of low order polynomials, but other natural choices are trigonometric, logarithmic, or exponential functions. 

With these ingredients, we seek a solution to
\begin{equation}\label{eq:sindy}
    \dot{\mathbf{X}}=\Theta(\mathbf{X}) \Xi
\end{equation}
for $\Xi=\left[\begin{array}{llll}\xi^{(1)} & \xi^{(2)} & \dots & \xi^{(d)}\end{array}\right]$ to learn the dynamics of each state variable $\dot{\mathbf{X}}_{i}=\mathbf{f}_{i}(\mathbf{X})=\Theta\left(\mathbf{X}\right) \mathbf{\xi}^{(i)}$. Framed as an optimization problem, the standard objective is to
\begin{equation}\label{eq:sindy_opt}
    \min_{\Xi} \|\mathbf{\dot{X}} - \Theta(\mathbf{X}) \Xi\|^2 + \lambda R(\Xi)
\end{equation}
 where $R(\cdot)$ is a sparsity promoting regularization function which may also include an $l_2$ ridge regularization term to improve the conditioning and add robustness \cite{bertsimas2018characterization}.

\paragraph{SINDy for PDEs}
This core framework can be further extended to the automatic discovery of PDEs by including partial derivatives in the candidate library \cite{schaeffer2017learning, rudy2017data}. Concretely, spatiotemporal data of $m$ spatial locations measured over $n$ time slices is arranged into a length $mn$ column vector ${\bf{U}}$. Then a candidate library $\Theta({\bf U}) \in \mathbb{R}^{mn \times D}$ is constructed as before except here we consider functions of both the system state and system spatial derivatives (which have to be numerically approximated). That is, in addition to library functions like $\mathbf{U}^2$, we also might include $\mathbf{U}\mathbf{U}_x$, and potentially higher order derivatives like $\mathbf{U}_{xxx}$. Finally, as before, we seek coefficients ${\Xi}$ which accurately reconstruct the temporal dynamics $\mathbf{U}_t = \Theta(\mathbf{U}) {\Xi}$.

\paragraph{Weak Form}
One core drawback with SINDy, especially when applied to PDEs or noisy data, is the need to numerically estimate derivatives because numerical differentiation compounds any noise present in the underlying measurement data. When differentiating multiple times, as is necessary for higher order PDEs, the estimates can become unusable, even while using more robust differentiation techniques like smoothed finite difference or polynomial interpolation.

This drawback motivates the weak form of SINDy \cite{schaeffer_sparse_2017, messenger2021weak, reinbold_using_2020} (which also generalizes to the PDE case \cite{gurevich2019robust, messenger_weak_2021}), where both sides of Equation~\ref{eq:sindy} are integrated over a random collection of $K$ temporal subdomains (spatiotemporal for PDEs). That is, for a random subdomain $\Omega_k$, a candidate library function $\mathbf{f}_i$, and a weight vector $\mathbf{w}$, we compute
\begin{equation} \label{eq:weak}
    q_i^k = \int_{\Omega_k} \mathbf{w}^T \mathbf{f}_i d\Omega
\end{equation} 
for every library term and subdomain. Each of the elements are then organized into a data matrix ${\bf Q} \in \mathbb{R}^{K \times D}$. By also integrating the left hand side of Equation~\ref{eq:system} over the same subdomains,  we get a linear system $\bf{q}_0 = {\bf Q} \Xi$ amenable to sparse regression without needing to differentiate noisy data. 
% We refer the reader to \cite{reinbold_using_2020, messenger2020weak} for more details on how to choose suitable weight vectors, wh

\subsection{Mixed-Integer Sparse Regression}
The seminal paper on best subset selection using MIO \cite{bertsimas2016best} proposed two primal formulations for sparse regression. Both of these formulations use binary variables to encode the support of the coefficients, one using big-M constraints, and the other using type-1 specially ordered sets (SOS-1). Because the system identification problem is coordinate separable, we can fit each dimension independently, resulting in $d$ smaller subproblems which can be solved directly using these techniques.

For system dimension $j$, with user-specified target sparsity $k_j$, the big-M formulation with ridge regularization for the SINDy problem is
\begin{align}
    \min_{\xi, z} \quad & \|\dot{X_j} - \Theta(X) \xi\|_2^2  + \lambda\|\xi\|_2^2\\
    \text{s.t.} \quad & M^\ell_i z_i \leq \xi_i \leq M^U_i z_i & i=1, \dots D \\
    & \sum_{i=1}^D z_i \leq k_j \\
    & \xi_i \in \mathbb{R}, \quad z_i \in \{0, 1\} & i = 1, \dots, D
\end{align}
where $M^\ell_i, \ M^U_i$ are lower and upper bounds on the coefficients. This formulation is solved for each $j \in [1, d]$, where we simply stack the coefficient vectors to recover the full system dynamics $\Xi=\left[\begin{array}{llll}\xi^{(1)} & \xi^{(2)} & \dots & \xi^{(d)}\end{array}\right]$. The theory and practice of solving MIO problems is deep, so we refer the interested reader to a standard text on the subject \cite{bertsimas2005optimization}. We rely on modern optimization solvers such as Gurobi \cite{gurobi} or CPLEX \cite{cplex2009v12} to both solve the problem and present a certificate of optimality. 

The effectiveness of big-M modeling relies on the tightness of coefficient bounds as otherwise the linear relaxations are too weak to efficiently prune the branch-and-bound tree. \cite{bertsimas2016best} derive a number of ways to obtain such bounds, however these approaches can add significant overhead to the solution times, and don't generalize well in the presence of arbitrary constraints. This motivates a nonlinear approach which circumvents the need to calculate these $2D$ different bounds. \cite{bertsimas2016best} also proposed adding the cardinality constraint via type-1 specially ordered sets (SOS-1) \cite{bertsimas2005optimization}. An SOS-1 constraint on a set of variables enforces that no more than one variable within the set is nonzero. In this case
$$(1-z_i)\xi_i = 0 \iff \{\xi_i, 1 - z_i\} : \text{SOS-1}$$
correctly captures the support of $\xi$. By replacing the support constraint, we get
\begin{align}
    \min_{\xi, z} \quad & \|\dot{X_j} - \Theta(X) \xi\|_2^2 + \lambda\|\xi\|_2^2 \\
    \text{s.t.} \quad & \{\xi_i, 1 - z_i\} : \text{SOS-1} & i=1, \dots D \\
    & \sum_{i=1}^D z_i \leq k_j \\
    & \xi_i \in \mathbb{R}, \quad z_i \in \{0, 1\} & i = 1, \dots, D.
\end{align}

More explicitly, the main objective term is
\begin{equation}
    \|\dot{X_j} - \Theta(X) \xi\|_2^2 = \xi^T \Theta(X)^T \Theta(X) \xi - 2 \langle \Theta(X)^T \dot{X_j}, \xi\rangle + \dot{X_j}^T \dot{X_i}.
\end{equation}
If we remove the constant term $\dot{X_j}^T \dot{X_j}$ and add the regularization term we get our final objective
\begin{align}
     \xi^T \Theta(X)^T \Theta(X) \xi - 2 \langle \Theta(X)^T \dot{X_j}, \ \xi \rangle + \lambda \xi^T \xi \\
     = \xi^T \left(\Theta(X)^T \Theta(X) + \lambda I\right) \xi - 2 \langle \Theta(X)^T \dot{X_j}, \ \xi \rangle .
\end{align}

For a problem with $n$ temporal observations and library size $D$, our final formulation has $D$ continuous variables, $D$ binary variables, $D$ corresponding SOS-1 constraints, and one knapsack constraint. The objective has $\frac{D (D+1)}{2}$ quadratic terms and $D$ linear terms. Notably, the formulation 
size is independent of $n$ which yields very favorable scaling properties as we will discuss in Section~\ref{sec:comp_eff}.

A question that naturally arises is whether to enforce sparsity as a hard constraint or promote sparsity as an objective penalty. While it may be tempting to use a penalty term and let the model balance sparsity, it is known that constrained problems enjoy more favorable statistical properties \cite{shen2013constrained}. In particular, while an optimal solution to the sparsity regularized problem is always obtainable by the constrained problem, the converse is not true in general (see \cite{kreber2019cardinality} Section~2.2). This is especially true when the data matrix exhibits high multicollinearity  which is common in system discovery because the library terms are usually strongly correlated.

Finally, we note that we adopt the SOS-1 formulation because we found it to be the most numerically stable and most flexible in including other potential model desiderata (e.g., satisfaction of physical constraints). However, it is not the most scalable. Modern general purpose optimization solvers can only handle sparse regression problems with up to a few thousand SOS-1 constraints. This is in stark contrast to very recent tailored sparse regression solution techniques such as the outer approximation method \cite{bertsimas_scalable_2020}, coordinate-descent based branch-and-bound \cite{hazimeh_sparse_2021-1}, and the backbone method \cite{bertsimas2022backbone} which can scale to the high-dimensional regime with dimension $O(10^7)$. Given that even a six-dimensional system with a fifth-order polynomial library is only of dimension 462, the more stable and flexible general purpose solvers are preferable for our circumstances.

\subsection{Extensions} \label{subsec:extensions}
In many physical systems where something is known about the underlying physics, we can incorporate this knowledge as constraints on the model coefficients. However, these constraints generally apply to the system as a whole (e.g., conservation of energy), so it is no longer possible to fit one coordinate at a time. Therefore, we fit all coordinates jointly using objective
\begin{equation}\label{eq:joint_opt}
    \min_\xi \left|\left|\begin{bmatrix}
\dot{X_1} \\ \dot{X_2} \\ \vdots \\ \dot{X_d} 
\end{bmatrix} - 
\begin{bmatrix}
\Theta(X) & 0 & \dots & 0 \\
0 & \Theta(X) &  & 0 \\
\vdots & &\ddots & \vdots \\
0 & 0 & \dots & \Theta(X)
\end{bmatrix} \begin{bmatrix} \xi^{(1)} \\ \xi^{(2)} \\ \vdots \\ \xi^{(d)}\end{bmatrix} \right|\right|_2^2 = \min_\xi \sum_{i=1}^d \xi^{(i)T} \Theta(X)^T \Theta(X) \xi^{(i)} - 2  \langle \Theta(X)^T \dot{X_i},  \xi^{(i)} \rangle.
\end{equation}

Now in addition to the sparsity constraint, we can add arbitrary constraints $A\bar{\xi} \leq b$ where $\bar{\xi}$ is the vectorized coefficient matrix of length $Dd$. Of course, because we inherit the full generality of MIO, these constraints can be anything that modern optimization solvers can handle (e.g., linear, quadratic, semidefinite, equality, inequality).

While such a formulation increases the dimension of the regression problem by a factor of $d$ (which is potentially quite costly, both in terms of computational time and sample efficiency), it is sometimes more natural and offers several additional benefits. The first is based on the fact that the coordinates used in SINDy often represent the spatial modes of a high-dimensional discretized PDE simulation computed using a dimensionality reduction technique like proper orthogonal decomposition \cite{chatterjee2000introduction}. Each of these spatial modes have an associated energy $\lambda_i$ designated by their singular values. Consequently, the quality of the high-dimensional reconstruction does not depend uniformly on the accuracy of each of the individual spatial modes, but in proportion to their energies. In this setting, we can weight each inner term in the sum of the objective function~\ref{eq:joint_opt} by the energy of the respective spatial mode to recover the terms which maximize the quality of the full reconstruction, rather than the average dimension-wise reconstruction. Another advantage is instead of having to run parameter tuning on different values of $k$ for each system variable, we can specify and cross-validate one global value of the desired sparsity, and the model will automatically determine the correct level of sparsity per dimension. 

While not pursued here, there are many extensions that could be appropriate for more specific circumstances. For instance, it is possible to add lower and upper bounds on the magnitudes of coefficients, add more sophisticated conditional logic on the relationship between nonzero coefficients, put controls on the level of multicollinearity \cite{bertsimas_scalable_2020}, require coefficients be statistically significant \cite{bertsimas_scalable_2020}, automatically prune outliers \cite{thompson2022robust, champion2020unified}, and much more. In short, as such a general framework, MIO fully empowers the researcher to express a vast range of model desiderata and impose additional structure to aid the learning process. 

\section{Results}
We benchmark our approach on nine canonical dynamical systems across a wide variety of statistical regimes. Our analysis focuses on attributes that most differentiate SINDy from alternative techniques, and then illustrates how using optimal methods furthers these advantages. In particular, we study sample efficiency, robustness, constraint enforcement, and computational efficiency where our evaluation focuses on identifying correct sparse models.

All of our experiments are built off of the open source PySINDy library \cite{de_silva_pysindy_2020, Kaptanoglu2022}. We use a shared university cluster with heterogeneous hardware where individual trials are confined to one CPU core with sufficient RAM. Our implementation of MIO sparse regression utilizes Gurobi 9.5.0 \cite{gurobi} to optimize and prove optimality. We make all of our code, data, and results publicly available at our Github repository \url{https://github.com/wesg52/sindy_mio_paper}.

\subsection{Experimental Overview}
Most of our experiments follow the same high level structure. We vary a quantity of interest (e.g., data length, data noise) for 50 random initial conditions, each with additive Gaussian noise scaled to be a certain percentage of the $l_2$ norm of the training data. Then for each sampled trajectory, we split the data into a training and a validation segment, using the validation segment to select the hyperparameters (see Section~\ref{subsec:modelselec}) of each of the baseline algorithms (Section~\ref{subsec:algos}). We follow the standard practice of unbiasing the final model by refitting an unregularized least squares regression on the selected coefficients. This final model is then evaluated on a suite of metrics (Section~\ref{subsec:eval}), with a focus towards identifying the correct coefficient support.

While the specifics of our results are somewhat sensitive to the details of our experimental procedure, we take steps to ensure our conclusions are robust to such design choices. For instance, we use random initial conditions whereas many papers in the SINDy literature report results using a fixed initial condition. While this aids reproducibility, we found performance to be sensitive to initial conditions for many systems, especially in the low data limit. Additionally, for each experiment we test our approach on multiple systems, each of which raises different qualitative behavior, to better understand the factors which differentiate performance.
%where our results makes clear the significant variance in qualitative behavior.

\subsubsection{Model Selection}
\label{subsec:modelselec}

A critical component of learning parsimonious models is in choosing hyperparameters that appropriately tradeoff model fit with sparsity, since adding more degrees of freedom monotonically decreases insample error. We strike this balance by selecting parameters which minimize the Akaike information criterion (AIC) metric \cite{akaike1974new, mangan_model_2017}.

In our setting of sparse regression, for a learned model $\hat{\Xi}$, the corrected AIC is given by
\begin{equation}
    AIC_c = m \ln(RSS / m) + 2k + \frac{2(k+1)(k+2)}{m-k-2}
\end{equation}
where RSS is the residual sum of squared errors $\sum_{i=1}^n \sum_{j=1}^d (\dot{X} - \Theta(X) \hat{\Xi})_{ij}^2$, $m = n \times d$ is the total number of measurements, $k$ is the sparsity of the solution, and the last term is the correction for finite samples.

Unless otherwise noted, for every trial of every experiment discussed below, we run the following model selection procedure. Split the sampled trajectory into a train and a validation interval, typically the first 2/3 and the last 1/3 respectively. For each algorithm and choice of hyperparameters, train on the training split and compute the $AIC_c$ with respect to the validation data. The final model is the one which minimizes the $AIC_c$ metric on the validation data. Note, for algorithms which fit each dimension separately, we compute the $AIC_c$ dimensionwise and combine the best coefficients per dimension to create the final model.

\subsubsection{Algorithms}
\label{subsec:algos}
For sake of consistent comparison, we restrict our baselines to methods of the same class, that is, $l_0$ regularized or constrained sparse regression algorithms. This notably excludes $l_1$ regularized sparse regression methods like LASSO and its variants \cite{tibshirani1996regression, carderera_cindy_2021}, probabilistic methods, and deep learning based approaches. Specifically, we compare our MIO based approach (MIOSR) to four common optimizers in the SINDy literature: sequential threshold least squares (STLSQ) \cite{brunton_discovering_2016}, sparse relaxed regularized regression (SR3) \cite{champion2020unified}, stepwise sparse regression (SSR) \cite{boninsegna_sparse_2018}, and ensembling using STLSQ (E-STLSQ) \cite{fasel2022ensemble}.

For every experiment trial, we perform a grid search over the relevant hyperparameters of each algorithm to find the set which minimize the $AIC_c$ as described above. Each algorithm has a hyperparameter corresponding to regularization strength and sparsity promotion. MIOSR, STLSQ, SSR, and E-STLSQ all have an explicit ridge regression penalty while SR3 has a relaxation parameter $\nu$. For MIOSR and SSR, we tune the sparsity of each dimension, where we control the can control $k$ exactly. For (E-)STLSQ and SSR, we tune the threshold parameter which acts as a proxy for the true sparsity. Parameter ranges for each algorithm are included in the appendix.

For SSR, we use the greedy criterion of removing the smallest magnitude coefficient at each iteration. For SR3, we run until convergence up to a max 10000 iterations. For MIOSR, we set a timeout of 30 seconds per dimension to prove convergence. For E-STLSQ we use robust bagging (bragging) where we randomly sample time slices with replacement to train 50 different models, and then use the median of the coefficients as the final model.

% MIOSR easier to tune

% For hyperparameter selection, we use a separate validation trajectory with the same duration and sample rate as the training trajectory, but independent initial condition and noise. This is without loss of generality, as one could split a single trajectory in half, creating a train and validation trajectory with the same parameters and different initial conditions. However, for systems with initial transience followed by fixed points or limit cycles only validating on half of the system complicates model selection, so we use two trajectories to enable a consistent experimental design across all systems.

\subsubsection{Evaluation Metrics}
\label{subsec:eval}
To evaluate each algorithm, we focus on three common metrics: true positivity rate, normalized coefficient error, and root mean squared error (RMSE) of the estimated dynamics. Throughout, let $\Xi$ be the true dynamics of the system and $\hat{\Xi}$ be the estimated dynamics.

The true positivity rate measures the ability to identify the correct nonzero terms of the dynamics, and importantly, to not select superfluous terms. Specifically, the true positivity rate is the ratio of intersection over union of nonzero coefficients 
\begin{equation}
    \frac{|\{i : \Xi_i \neq 0\} \cap \{i : \hat{\Xi}_i \neq 0\}|}{|\{i : \Xi_i \neq 0\} \cup \{i : \hat{\Xi}_i \neq 0\}|},
\end{equation} or equivalently, the ratio of true coefficients identified to the combined number of true coefficients, false zero coefficients, and false nonzero coefficients identified.

The normalized coefficient error simply measures the normalized euclidean distance between the true coefficients and the learned coefficients 
\begin{equation}
    \frac{\|\Xi - \hat{\Xi}\|_2}{\|\Xi\|_2}.
\end{equation} This metric is less punitive that the true positivity rate because it mainly captures the difference in large coefficients, and minimally penalizes small nonzero terms.

Finally, to contextualize how the coefficient error actually impacts the quality of the recovered model, we also report the root mean squared error of the estimated derivatives on a clean test set. That is, for a testing trajectory of length $n$ we calculate 
\begin{equation}
    \sqrt{\frac{1}{nd} \sum_{i=1}^n \sum_{j=1}^d (\Theta(\mathbf{X}_{test})\Xi - \Theta(\mathbf{X}_{test})\hat{\Xi})_{ij}^2}.
\end{equation}
To test that the model truly generalizes, we independently sample 10 initial conditions, each run for 10 seconds, and take the average RMSE across each of these trajectories. Due to the chaotic nature of the systems we study, we only compare the derivatives and not the trajectories of the forward models, since initial errors will rapidly compound regardless of the accuracy of the underlying model.

In the results that follow, we will typically report the log RMSE and log $l_2$ coefficient errors, averaged over 50 trials. When calculating the mean and standard errors, we do so on the log of these statistics. This avoids one outlier from dragging up the mean by many orders of magnitude, and makes the standard errors symmetric in log space.

\subsection{Sample Efficiency}
\label{sec:sample_eff}
A key advantage of SINDy over deep learning techniques is the reduced data requirements to learn high quality sparse models. To test how MIOSOS extends this advantage, we study the effect of varying the length of the training trajectory on model recovery for three canonical dynamical systems studied in the SINDy literature: the Lorenz model \cite{lorenz1963deterministic}, the Hopf system \cite{marsden2012hopf}, and a triadic magnetohydrodynamical (MHD) model \cite{carbone1992relaxation}.

The Lorenz model is the classic example of a chaotic system and is given by
\begin{subequations}
    \begin{align}
        &\dot{x}=\sigma(y-x) \\
        &\dot{y}=x(\rho-z)-y \\
        &\dot{z}=x y-\beta z
    \end{align}
\end{subequations}
where we use the standard parameters $\sigma = 10, \ \beta = 8/3, \ \rho=28$. Another canonical system in the study of nonlinear dynamics is the Hopf system given by
\begin{subequations}
    \begin{align}
        &\dot{x}=\mu x+\omega y-A x\left(x^{2}+y^{2}\right) \\
        &\dot{y}=-\omega x+\mu y-A y\left(x^{2}+y^{2}\right)
    \end{align}
\end{subequations}
where we set $\mu = -0.05, \ \omega = A = 1$. Finally, we consider a simplified plasma model of a joint velocity and magnetic field with 0 fluid viscosity or resistivity

\begin{subequations}
    \begin{align}
        \dot{V}_1 & = 4\left(V_{2} V_{3}-B_{2} B_{3}\right) \\
        \dot{V}_2 & = -7\left(V_{1} V_{3}-B_{1} B_{3}\right) \\
        \dot{V}_3 & = 3\left(V_{1} V_{2}-B_{1} B_{2}\right) \\
        \dot{B}_1 & = 2\left(B_{3} V_{2}-V_{3} B_{2}\right) \\
        \dot{B}_2 & = 5\left(V_{3} B_{1}-B_{3} V_{1}\right) \\
        \dot{B}_3 & = 9\left(V_{1} B_{2}-B_{1} V_{2}\right).
    \end{align}
\end{subequations}

\begin{figure}
    \centering
    \includegraphics[width=\linewidth]{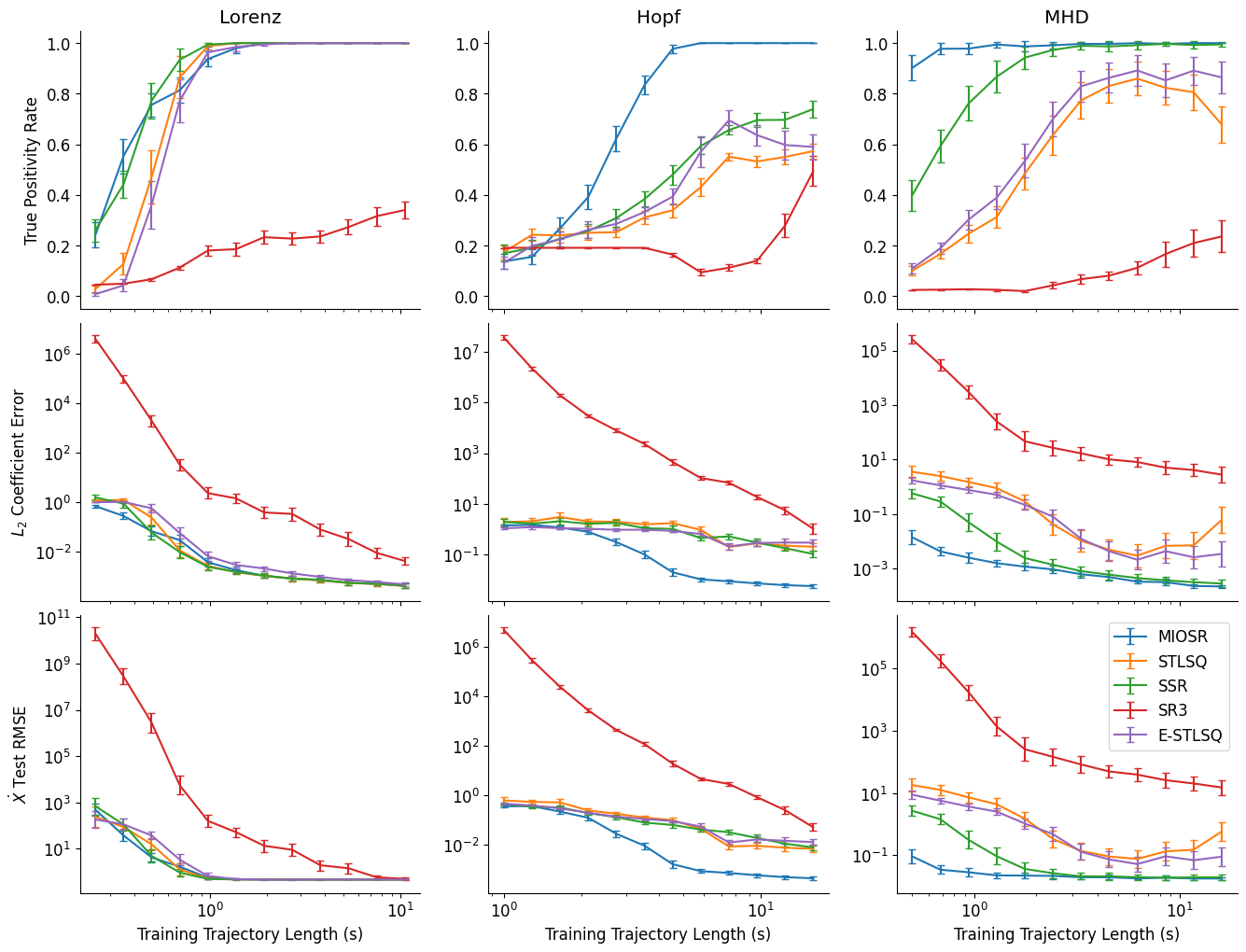}
    \caption{Performance comparison of sparse regression algorithms for the differential form of SINDy under varying amounts of training data for three different canonical systems: Lorenz, Hopf, and MHD. Results are averaged over 50 trials with added Gaussian noise of 0.2\%.}
    \label{fig:differential}
\end{figure}
Figure~\ref{fig:differential} depicts our results for each combination of algorithm and system for varying lengths of training data. For each system we sample trajectories with 0.002 second time granularity with $0.2$\% added Gaussian noise. In anticipating the assumption that exact optimization methods are not scalable, we use oversized libraries: 5th order polynomials for the Lorenz and Hopf systems and 3rd order polynomials for MHD. This yields libraries with dimension 56, 21, and 84 respectively.

The high level conclusion is that MIOSR consistently out performs all other methods by finding more accurate models with less data, and with less variance in the quality of the fit. While the gap is more muted in the Lorenz case, the stark differences in the Hopf and MHD systems surface two distinct scenarios where heuristics can fail: the presence of small coefficients and large libraries. For Hopf, with bifurcation parameter $\mu =-0.05$, thresholding techniques cannot select such a small coefficient in the presence of even modest noise. Indeed, the original SINDy paper used multiple independent trajectories to learn the Hopf dynamics, because the system quickly converges to a fixed point. In the low data limit for MHD, a larger 6 dimensional system,  there are many combinations of the large library of terms which fit the small training set well. With so many degrees of freedom, iterative methods break down by taking incorrect intermediate steps, but by taking a global view, MIOSR can identify the true model.

In this low data regime, some baselines completely fail, in particular SR3. This is perhaps unsurprising because SR3 fits all coordinates jointly and therefore is solving a more difficult, higher dimensional optimization problem. While sometimes the baselines methods achieve achieve comparable test RMSE, they often do so by overfitting as evidenced by the low true positivity rate. This largely defeats the purpose of SINDy in discovering robust, interpretable, and scientifically illuminating models.

Many of the aforementioned difficulties are partially ameliorated by using a smaller library. In the appendix, we perform the same experiment with ``tight'' libraries, those which don't include higher order polynomials than are necessary to express the system (Figure~\ref{fig:appendix_differential}). While MIOSR maintains a clear edge, the difference is less striking. Of course for novel systems, this information is not available \textit{a priori}, and therefore this represents the most idealized scenario.

% Use large library
% 
% SR3 works better when using a smaller library.

\subsection{Computational Efficiency}
\label{sec:comp_eff}

\begin{figure}
    \centering
    \includegraphics[width=\linewidth]{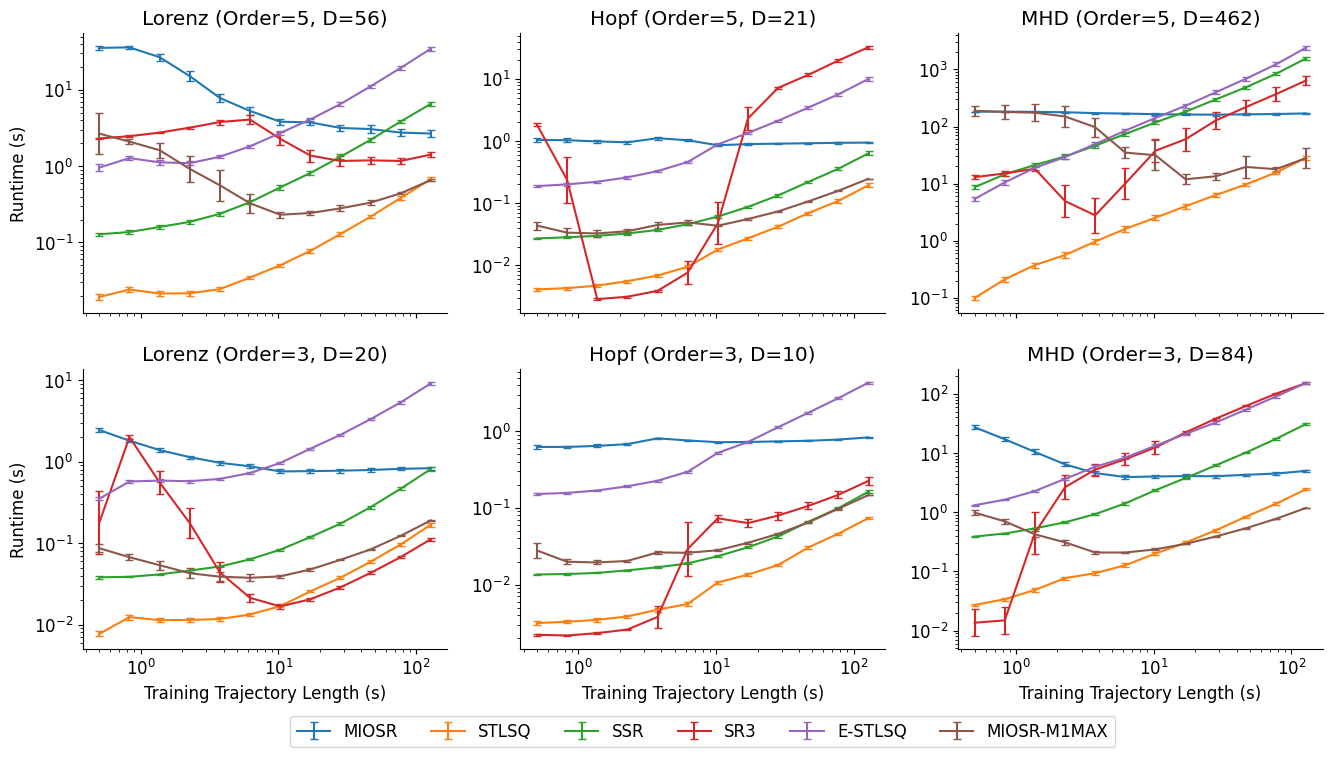}
    \caption{Comparison of sparse regression algorithm computational efficiency for the differential form of SINDy under varying amounts of training data for three different canonical systems: Lorenz, Hopf, and MHD. The top row uses a fifth order polynomial library for each of the three systems while the bottom row uses a third order polynomial library. Results are averaged over 50 trials with added Gaussian noise of 0.2\% }
    \label{fig:runtime}
\end{figure}

Nearly every SINDy paper contains a sentence that justifies the need for sparse regression heuristics by claiming that the feature selection problem is computationally intractable due to the combinatorial nature. To directly dispel this claim, we compare the wallclock computation time of each of the different algorithms for varying library sizes and amounts of training data (see Figure~\ref{fig:runtime}). Similar to the previous experiment, for each system, library size, and amount of data, we train each algorithm on 50 random trajectories with 0.2\% additive Gaussian noise with 0.002 sample frequency. We precompute the derivatives and library, so the reported times corresponds to just the regression time, not the whole SINDy pipeline. Unlike the previous experiment, we are not performing hyperparameter tuning, and instead use appropriate defaults learned above.

Several points deserve elaboration. Perhaps most surprising to those unfamiliar with MIO based machine learning, is that MIOSR is often faster as the amount of data increases, sometimes significantly \cite{bertsimas2020sparse}. This is partially due to the fact that the final optimization problem has no dependence on $n$, as we only require a one time $n \times D$ matrix multiplication to initially construct the objective value coefficients. On the other hand, for small $n$, there are more ways to fit the data, so the bounds in the branch-and-bound tree are weaker, necessitating more node exploration. Combining these results with those from Section~\ref{sec:sample_eff}, we conclude that either MIOSR takes a comparable amount of time while achieving the same accuracy, or it takes longer but the extra computational cost buys extra statistical performance. That is, regardless of data size, the cost of MIOSR is justified.

Another critical point is the dramatic speed up associated with utilizing more powerful hardware and parallelization. When run on a high end laptop (2021 Macbook Pro with M1 Max chip), as opposed to a single core of a cluster node, MIOSR can be up to 100x faster, while all other methods only improve by a few percent (see Figure~\ref{fig:appendix_runtime}). This is because MIO is a highly parallelizable technology, as multiple threads can explore, expand, and prune different branches of the branch-and-bound tree in parallel. Therefore, one can simply allocate additional compute to achieve increasing levels of performance, and the time gap between optimal methods and heuristics will continue to close as more powerful MIO software and parallel hardware emerge in the future.

In addition to hardware, there are several other factors which understate how efficient MIOSR can be relative to the results reported in Figure~\ref{fig:runtime}. The first is that MIO requires less hyperparameter tuning than other methods because we are directly tuning the sparsity and not a proxy threshold. Second, for especially difficult or large problems, one can use warm starts from heuristic methods to get good initial solutions, or reuse solutions and models from previous steps in the hyperparameter search (which also avoids reconstructing the full optimization model). Finally, much of the computational effort is dedicated toward proving optimality by improving the dual lower bound \cite{bertsimas2020sparse}. Therefore, one could set a short timeout on the solver to get what is likely an optimal solution, but give up on the optimality guarantee if so desired.

% precompute derivatives and library functions
% timeout, proving the dual
% MIO scales with CPU
% phase transitions
% warmstarts
% This is only one trial, also have to consider how much tuning is required 

\subsection{Physical Constraints}
\label{sec:physcon}
A central goal within scientific machine learning is to incorporate existing physical knowledge into the models, both to aid the learning process and to ensure that physically plausible models are learned. To illustrate the improved capability of MIOSR in service of this goal, we replicate the experiment performed by \cite{champion2020unified} on the two-dimensional Duffing system.

To briefly describe their setup, the 2D Duffing system is both a Hamiltonian system and a gradient system. These properties induce constraints on the coefficients because each individual governing equation must be a partial derivative of a Hamiltonian or potential function. The full system is described by 

\begin{subequations}
    \begin{align}
        \dot{x} &=X \\
        \dot{y} &=Y \\
        \dot{X} &=-\frac{\partial}{\partial x} V(x, y) \\
        \dot{Y} &=-\frac{\partial}{\partial y} V(x, y)
    \end{align}
\end{subequations}
where $x, \ y$ give the spatial position, $X, \ Y$ give the momentum, and the potential function is 
\begin{equation}
    V(x, y) = -\frac{\omega}{2}(x^2 + y^2) + \frac{\alpha}{4} (x^2 + y^2)^2.
\end{equation}
We use SINDy to just fit the spatial coordinates and set $\omega=-2$ and $\alpha=0.1$. We refer the interested reader to \cite{champion2020unified} for the detailed derivation of the constraints, but for our purposes, the relevant fact is simply that the potential function imposes a set of equality constraints on $\Xi$. We can then use a vectorized representation of the coefficients $\bar{\xi}$, and add $A \bar{\xi} = b$ to the MIO model where $A, \ b$ are based on the partial derivatives of $V(x, y)$. Because these constraints extend between dimensions, we use the joint formulation \ref{eq:joint_opt} to fit $x$ and $y$ simultaneously.

\begin{figure}
    \centering
    \includegraphics[width=.95\linewidth]{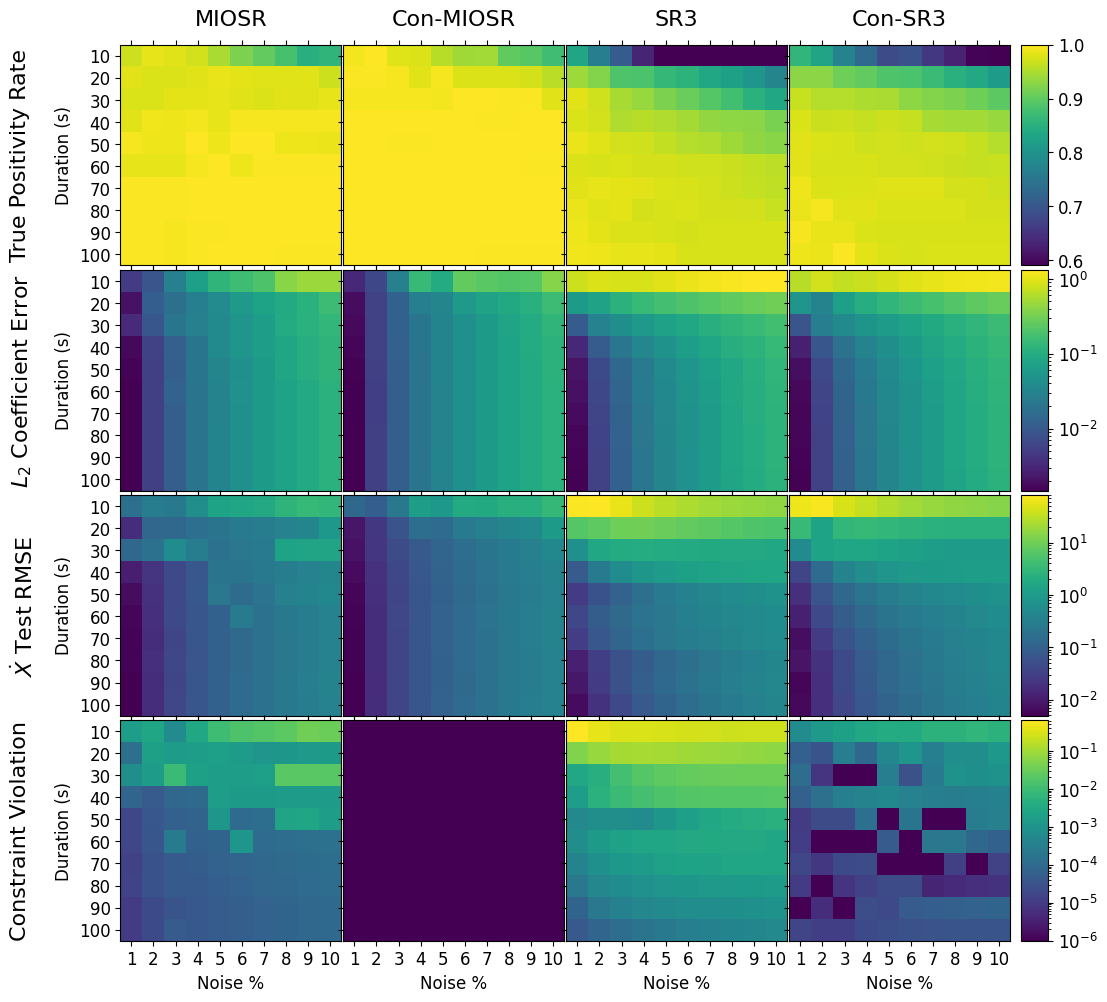}
    \caption{Performance comparison of constrained versus unconstrained sparse regression for the differential form of SINDy under varying amounts of training data and noise for the 2D Duffing system. Results are averaged over 50 trials with random initial conditions.}
    \label{fig:constraint}
\end{figure}
As in previous experiments, we sample 50 independent trajectories with random initial conditions, for every combination of training duration and noise depicted in Figure~\ref{fig:constraint}. For every trajectory, we run MIOSR with and without constraints, as well as SR3 with and without constraints, as it is the only baseline which naturally accommodates constraints. To stay consistent with \cite{champion2020unified}, we use a third order polynomial library with a sample rate of 0.01 seconds. Additionally, we do not unbias the coefficients after feature selection and report the average constraint violation $\frac{1}{c}\|A \hat{\xi} - b \|_1$ where $c$ is the number of constraints.

The immediate takeaways from Figure~\ref{fig:constraint} are that adding constraints help both algorithms, especially with limited data, but that even without constraints, MIOSR is more accurate. This is in slight contrast to the findings of \cite{champion2020unified} where ``the
constrained and unconstrained models have nearly identical
[$R^2$] scores at all noise levels.'' However, this is because we study a more difficult statistical regime, one with less data and with coefficients of different magnitudes. In particular, they used training sets that included 20 independent trajectories, hence the constraints did not add additional information. 

Another drawback of SR3 is that it enforces the constraints on a set of relaxed coefficients, hence the constraints are not strict and there exist nontrivial violations. This occurs most frequently in training regimes with less data and more noise. Unfortunately, these are precisely the regimes where constraints are most useful. MIOSR, in contrast, always satisfies constraints up to solver numerical precision, which can be adjusted or relaxed as desired. Finally, we observe that adding constraints does not add to the solution time of MIOSR, so the runtimes are comparable to those reported in Section~\ref{sec:comp_eff}, while constraining SR3 increases the runtime by about 30\%.

\subsection{Robustness}
\label{sec:robustness}
\begin{figure}
    \centering
    \includegraphics[width=\linewidth]{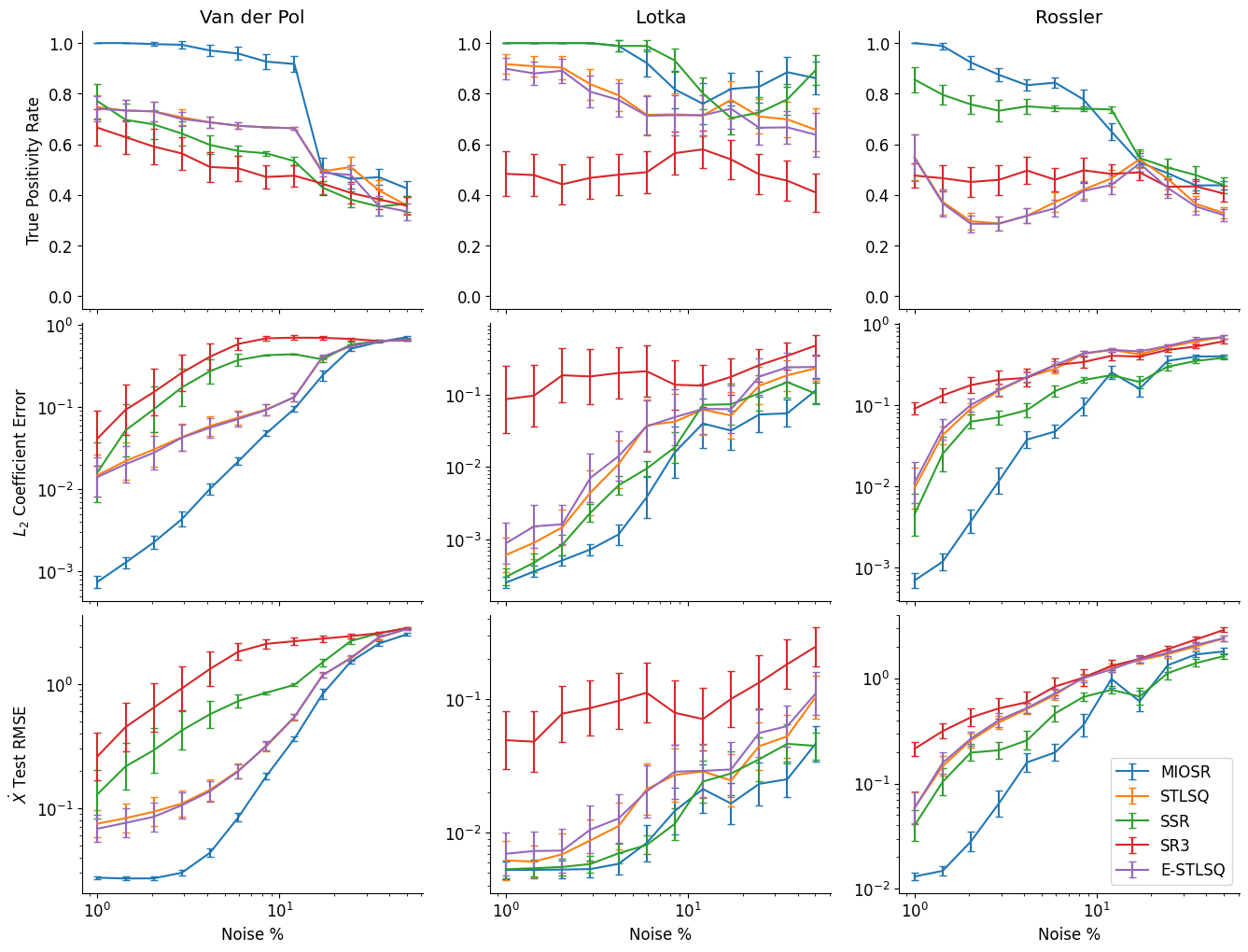}
    \caption{Performance comparison of sparse regression algorithms for the integral form of SINDy under varying amounts of added Gaussian noise for three different canonical systems: Van der Pol, Lotka, and Rossler. Results are averaged over 50 trials, each
    with 50 seconds of training data.}
    \label{fig:integral}
\end{figure}
For the remainder of our experiments, we study system recovery under substantial noise, and therefore, we use the weak form of SINDy, where the data matrix is given by Eq.~\ref{eq:weak}. We first study robust recovery of three canonical ODEs: the Van der Pol oscillator \cite{van1926lxxxviii}, Lotka-Volterra equations \cite{lotka1925elements}, and the R\"{o}ssler system \cite{rossler1976equation}.

The Van der Pol system is given by
\begin{subequations}
    \begin{align}
      \dot{x} & = y \\
      \dot{y} & = \mu (1 - x^2) y - x
    \end{align}
\end{subequations}
where we use $\mu=3$. The Lotka-Volterra equations, sometimes also known as the predator-prey equations given their origin in modeling wildlife populations, are given by
\begin{subequations}
    \begin{align}
      \dot{x} & = p_1 x - p_2xy \\
      \dot{y} & = p_2 xy - 2 p_1 y
    \end{align}
\end{subequations}
where we set $p_1 = 1$ and $p_2 = 10$. Finally the R\"{o}ssler system is
\begin{subequations}
    \begin{align}
      \dot{x} & = -y - z \\
      \dot{y} & = x + ay \\
      \dot{z} & = b - cz + xz
    \end{align}
\end{subequations}
where we have $a=b=0.2$ and $c=5.7$.

For all three systems, we use a third order polynomial library with 50 seconds of training data with time intervals of 0.002 seconds (i.e., 25000 total time steps). Our weak libraries are composed of 2400 spatial domains each with 400 points per domain. To validate weak models on noisy data, one can either use the weak form of the validation set, or try to differentiate the validation data with more aggressive smoothing. We observe that the weak form of the validation data yields a less reliable tuning signal, partially because the weak form uses randomized domains, and leads to less sparse models. However, the numerical derivative also becomes increasingly unreliable with more noise. Therefore, under 15\% noise we use the a smoothed derivative (with window size 21) for validation, and use the weak form of the validation data above 15\% additive noise.

%validation delicate
% ensemble library

Figure~\ref{fig:integral} depicts our results. The general trend is that for low to medium amounts of noise, MIOSR is significantly more accurate, often perfectly recovering the underlying model. Even with just 1\% noise, the baselines struggle to identify the true model coefficients. However, unlike in Section~\ref{sec:sample_eff}, the baselines are mostly identifying all of the correct coefficients, but find small false positive coefficients that are fitting noise.

At very high levels of noise, MIOSR starts to break down, converging to the approximate performance of heuristic methods. While more elaborate ensembling could help, we believe better data preparation is likely a more effective way to learn accurate models, especially with MIOSR. Examples include applying more aggressive smoothing, expanding the number of domains or points per domain in the weak form, or utilizing tailored differentiation techniques \cite{van2020numerical}. Beyond data preparation, there is also more recent work on more sophisticated iterative schemes to prune the library by regressing on Fourier transforms \cite{delahunt2021toolkit}, which could benefit from utilizing optimal methods.

\subsection{PDEs}

\begin{figure}
    \centering
    \includegraphics[width=\linewidth]{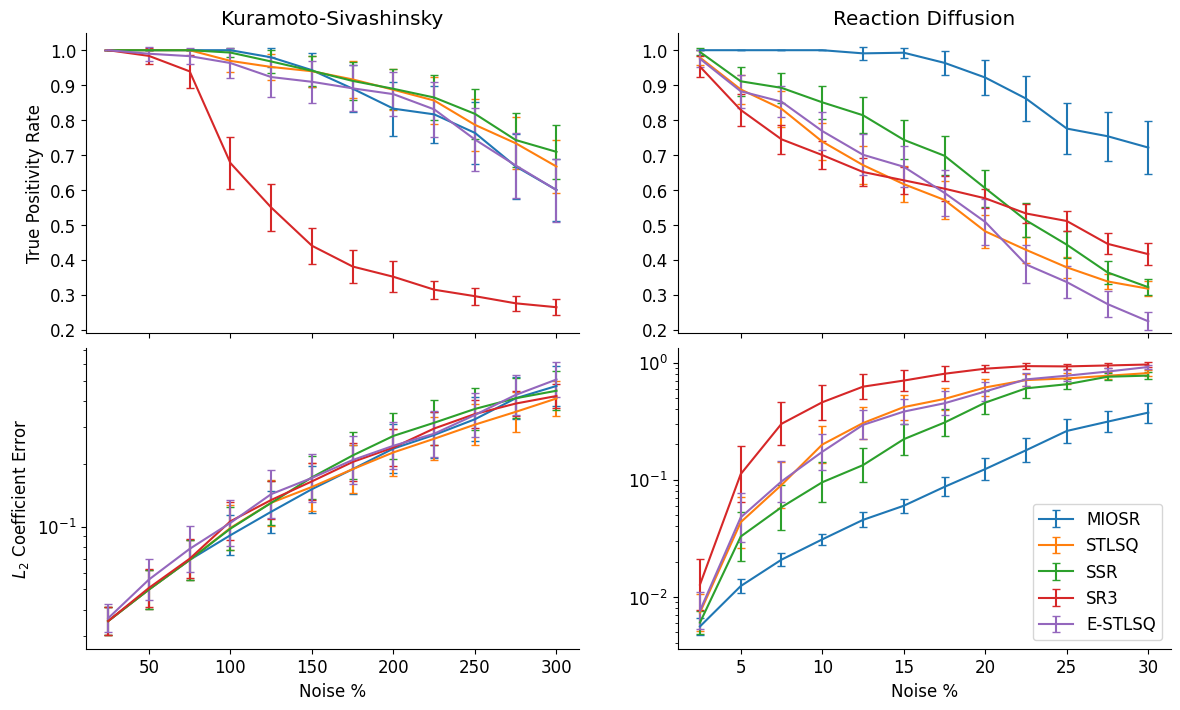}
    \caption{Best PDE model found by sparse regression algorithms for the integral form of SINDy under varying amounts of Gaussian noise for two different canonical PDE systems: Kuramoto-Sivashinsky and reaction diffusion. Results are averaged over 50 trials with random initial conditions.}
    \label{fig:pde}
\end{figure}
For our last experiment, we study the recovery of PDEs under substantial measurement noise. This is likely the regime most relevant to advancing modern science and engineering practice. In particular, we study the one-dimensional Kuramoto-Sivashinsky equation \cite{sivashinsky1977nonlinear}, an early model of laminar flame fronts, and the two-dimensional reaction diffusion system, a ubiquitous model in chemistry. The Kuramoto-Sivashinsky in spatial dimension $x$ and time $t$ is given by
\begin{equation}
    u_t = -uu_x - u_{xx} - u_{xxxx}
\end{equation}
where the notation $u_{xx}$ denotes the second partial derivative with respect to $x$. We study the 2D reaction diffusion system given by
\begin{subequations}
\begin{align}
    u_t = & \frac{1}{10} u_{xx} + \frac{1}{10} u_{yy} + u - uv^2 - u^3 + v^3 + u^2 v \\
    v_t = & \frac{1}{10} v_{xx} + \frac{1}{10} v_{yy} + v - uv^2 - u^3 - v^3 - u^2v.
\end{align}
\end{subequations}
For both systems we use weak third order polynomial libraries, with up to fourth order derivatives for Kuramoto-Sivashinsky and second order derivatives for reaction diffusion, yielding library dimensions of 19 and 109 respectively. For Kuramoto-Sivashinsky, our weak library is composed of 200 spatiotemporal domains, each with 50 points, sampled 10 times a second for 25 seconds on a periodic domain with 1024 spatial points. For reaction diffusion, our weak library is composed of 400 spatiotemporal domains, each with 36 points, sampled 50 times a second for 5 seconds on a $256 \times 256$ periodic grid.

Unlike the previous experiments, we do not perform model tuning and selection using AIC, due to the computational cost of fitting a large number of PDEs. Instead, we perform an achievability analysis, loosely inspired by \cite{maddu_stability_2019}, where we choose the algorithm parameters knowing the dynamics, to determine if it is even possible to learn a correct model given an appropriate model selection method. In particular, for MIOSR, we set the sparsity constraint to be the actual dimensionwise sparsity; for STLSQ, we try several thresholds slightly below the actual smallest coefficient; for E-STLSQ, we use library ensembling \cite{fasel2022ensemble} and take the $k$ terms with the highest inclusion probability, where $k$ is the true sparsity.

Figure \ref{fig:pde} depicts our results for PDE learning based on 50 trials with random initial conditions and varying amounts of additive Gaussian noise. These results further underscore how MIOSR has a substantial edge over heuristic methods when using larger libraries or when the system coefficients are of different orders of magnitude; it also further illustrates the converse, that problems with smaller libraries and similar magnitude coefficients do not benefit from exact methods because the heuristics converge to the correct solution. Regardless of optimizer, we see how effective the weak SINDy framework can be in identifying noisy systems, with Kuramoto-Sivashinsky often being recovered even with 300\% measurement noise.

% We note that reaction diffusion was the only system where our default timeout of 30 seconds per dimension was not sufficient to learn optimal solutions.

% RD only systems where timeout mattered

\section{Conclusion}
%contain truth and only the truth
%fits into any other sindy advancements 

In this work, we demonstrate the superior performance of mixed-integer optimization in learning sparse nonlinear dynamics as compared with popular heuristic methods. The biggest advantage is in finding models which are as simple as possible, but not simpler---a model which learns the truth and only the truth. In addition to more accurate support recovery, MIO sparse regression is capable of incorporating a huge range of additional model structure as auxiliary objective terms or constraints, while solving the underlying optimization problem to provable optimality.

Contrary to the predictions of complexity theory, MIOSR is highly tractable, and can actually be faster than heuristics for large amounts of data. Indeed, MIOSR runs slower when the regression is harder, that is, when the sample size is small, signal-to-noise ratio is low, or the coefficients span multiple orders of magnitude. However, this is exactly where heuristic methods perform poorly, so MIOSR requires more time when it improves upon heuristic methods while being comparable in running time when the dynamics are more easily recoverable. Given the practicality of the approach, and the theoretical guarantees, we see no reason why MIOSR should not be the default choice of optimizer for real applications.

Due to the modularity of the SINDy framework, MIOSR is compatible with other methodological advancements concerning data preprocessing, library construction, numerical differentiation, and outer loop algorithms. We restricted our study of these extensions to the weak form and PDE learning, but we expect MIOSR to offer similar benefits to other variants like control \cite{brunton2016sparse} or identifying implicit equations \cite{kaheman_sindy-pi_2020}. We hope domain experts find use for the additional modeling and statistical power afforded by MIO. We believe this is an exciting development that advances the state-of-the-art in system discovery.

%Bibliography
\bibliographystyle{unsrt}  
\bibliography{latex, SINDY}  

\appendix
\section{Additional Experiment Details}
Here we record in greater detail the set of initial conditions we use for each system, the parameter ranges in tuning the various algorithms, and other relevant implementation details. Additionally, in our raw results files made available on Github, we include the initial condition, random seed, and chosen hyperparameters for every trial in every experiment.

\subsection{Sample Efficiency}
For all systems and algorithms we use smoothed finite difference differentiation with a smoothing window length of 9. We use the same regularization grid for all systems. For MIOSR, STLSQ, and SSR we tune over the regularization strength $\alpha \in \{0, 10^{-5}, 10^{-3}, 10^{-2}, 0.05, 0.2\}$. For E-STLSQ we use the best $\alpha$ for STLSQ. For SR3 we tune relaxation parameter $\nu \in \{\frac{1}{30}, \frac{1}{10}, \frac{1}{3}, 1, \frac{10}{3} \}$. For thresholds, we try to tailor the range based on the system to give the best shot at finding a sparse model (since the heuristics are quite sensitive to the threshold). We choose 50 values uniformly in log space. That is, $10^a$ for $a \in [b:c:d]$ where $d=50$ values equally spaced on the interval $[b, c]$. In particular we use $[-2:1:50]$, $[-2:0:50]$, and $[-1.5:1.5:50]$ for Lorenz, Hopf, and MHD, respectively (where we increase the range by 0.5 for SR3 since it does use a hard threshold). For MIOSR, we tune the sparsity $k$ for each dimension over integers $k \in [1, 5]$, and SSR by nature fits a model at every level of sparsity between one and the full library size.

Regarding initial conditions, we sample uniformly from a specified volume. For Hopf, we sample in polar coordinates: a radius uniformly at random between 0.75 and 1.25 and an angle at random from 0 to $2\pi$ radians. For Lorenz, we sample $x, y \in [-5, 5]$ and $z \in [10, 40$. For MHD we sample every coordinate independently from $[-1.5, 1.5]$. We use the same sampling strategy for the runtime experiment.
\subsection{Physical Constraints}
For both algorithms we use smoothed finite difference differentiation with a smoothing window length of 21 to accommodate the noisier data. As before, for SR3 we tune over $\nu \in \{\frac{1}{30}, \frac{1}{10}, \frac{1}{3}, 1, \frac{10}{3} \}$ and 50 thresholds $\lambda=10^a, \ a \in [-3: 0: 20]$. For MIOSR, we tune over a global sparsity constraint $k \in [2, 10]$, and regularizer in $\{0.0001, 0.001, 0.01\}$. As in \cite{champion2020unified}, we sample initial conditions uniformly for each dimension in $[-\pi, \pi]$.
\subsection{Robustness}
For all weak form experiments we normalize the data matrix to have unit column norm. We use the same values of $\alpha$ and $\nu$ as before. Again, in an effort to get the best baseline model, we tailor the thresholds to the system and check that all chosen thresholds fall in the range we tune over. For Van der Pol, Lotka, and Rossler respectively we tune the threshold $\lambda$ over $2^a \ a \in [-1:5:50]$, $[-3: 4: 50]$ and $ [2: 6: 50]$ for STLSQ. For E-STLSQ we use the same regularization that was optimal for STLSQ.

For Van der Pol, we sample initial conditions from the box $x \in [-1, 1], \ y \in [-\mu, \mu]$ where we use $\mu=3$. For Lotka, we sample initial conditions from the box $x, y \in [0, 1]$. For Rossler, we sample uniformly from the a canonical trajectory with initial condition $(5, 3, 0)$ and add 10\% Gaussian noise. For Rossler, we additionally take the absolute value of the $z$ coordinate to preclude unstable trajectories.

\subsection{PDEs}

For achievability analysis, we don't need to tune the sparsity of MIOSR since we can simply use the true sparsity and check if the correct model is learned. However, for (E-)STLSQ the optimal threshold does change because we add substantial noise to the data. Therefore we still train for thresholds in $\lambda \in [0.4, 2.0, 20]$ and $\lambda \in [0.04, 0.16, 20]$ for Kuramoto Sivashinsky and reaction diffusion respectively.

The initial conditions for Kuramoto Sivashinsky are sampled as $\frac{1}{Z}(\cos(x + r_0) + \sin(4r_1 x))$ where $r_0, r_1 \in [0, 1]$ are sampled uniformly at random, $x$ is the 1D mesh on $[0, 2\pi]$ with 1024 grid points, and $Z$ is the normalization factor $\|\cos(x + r_0) + \sin(4r_1 x))\|_\infty$. For reaction diffusion we use a spiral initial condition that is randomly rotated and slightly expanded or contracted. In particular, for $X, Y$ representing the $256\times 256$ spatial mesh
$$u_0 = \tanh((X^2 + Y^2)^{1/2}) \cos(
                (\text{angle}(X + iY) + o)
                - (s(X^2 + Y^2)^{1/2})
            )$$
$s \in [0.95, 1.05]$ and $o \in [0, 2\pi]$ both sampled uniformly at random. $v_0$ is the same but with $\sin$ instead of $\cos$.

% what exactly was tuned for each algorithm, and the parameter ranges
% initial conditions for systems
\section{Additional Results}
\begin{figure}[h]
    \centering
    \includegraphics[width=\linewidth]{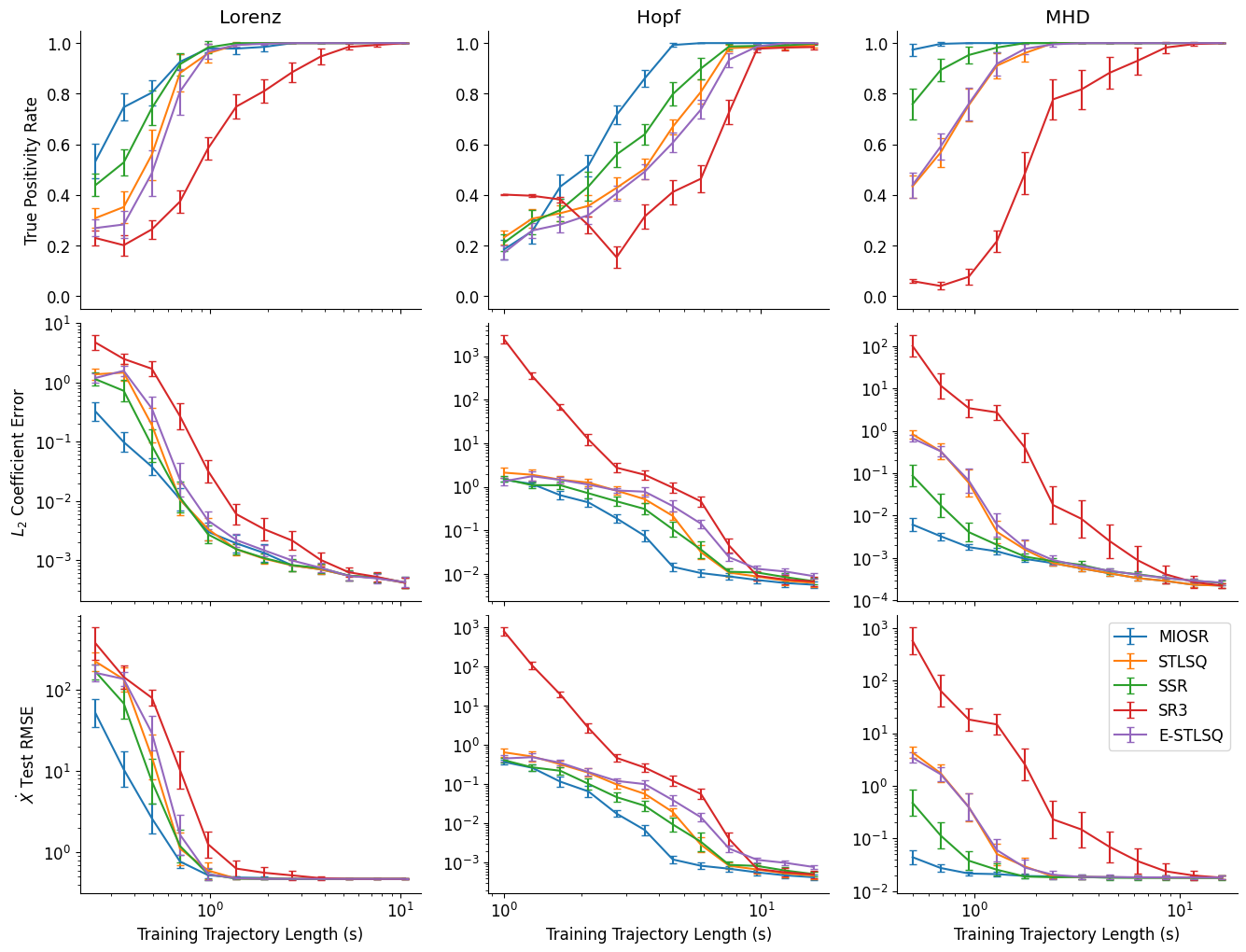}
    \caption{Performance comparison of sparse regression algorithms for the differential form of SINDy using minimal polynomial libraries under varying amounts of training data for three different canonical systems: Lorenz, Hopf, and MHD. Results are averaged over 50 trials with added Gaussian noise of 0.2\%.}
    \label{fig:appendix_differential}
\end{figure}% Local runtime

\begin{figure}
    \centering
    \includegraphics[width=\linewidth]{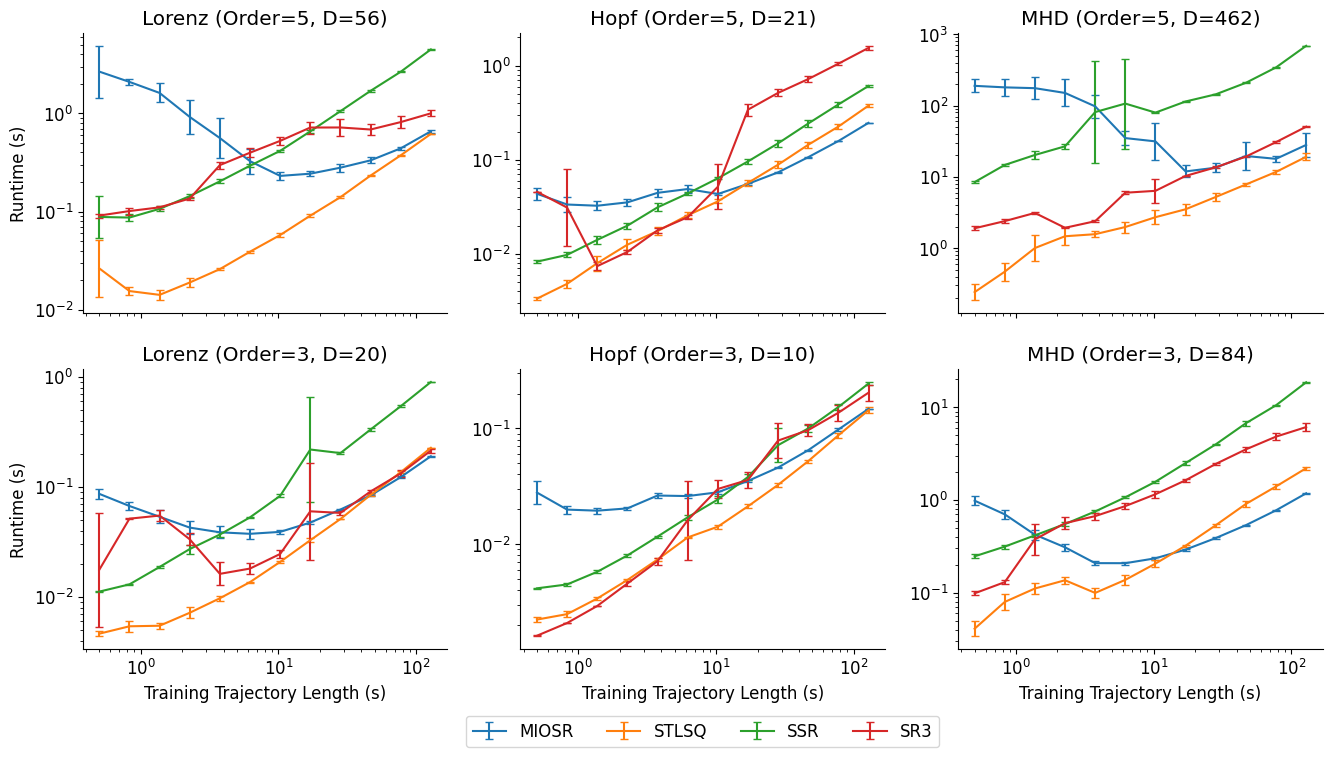}
    \caption{Comparison of sparse regression algorithm computational efficiency, when executed on a 2021 Macbook Pro, for the differential form of SINDy under varying amounts of training data for three different canonical systems: Lorenz, Hopf, and MHD. The top row uses a fifth order polynomial library for each of the three systems while the bottom row uses a third order polynomial library. Results are averaged over 5 trials with added Gaussian noise of 0.2\% }
    \label{fig:appendix_runtime}
\end{figure}

\end{document}